\documentclass[journal]{IEEEtran}

\usepackage[utf8]{inputenc}
\usepackage[T1]{fontenc}
\usepackage{amsmath,amssymb}
\usepackage{algorithm}
\usepackage{algpseudocode}
\usepackage{booktabs}
\usepackage{graphicx}
\graphicspath{{figures/}}
\usepackage{multirow}
\usepackage[table]{xcolor}
\usepackage{tikz}
\usetikzlibrary{arrows.meta,positioning,calc,shapes.geometric,fit,decorations.pathreplacing}
\usepackage{cite}
\usepackage{hyperref}
\hypersetup{colorlinks=true,linkcolor=blue,citecolor=blue,urlcolor=blue,filecolor=blue}
\usepackage[capitalise]{cleveref}
\crefname{figure}{Fig.}{Figs.}
\Crefname{figure}{Fig.}{Figs.}

\usepackage{fancyhdr}

\fancypagestyle{firstpage}
{
    \fancyhead[L]{\footnotesize Personal use of this material is permitted.  Permission must be obtained for all other uses, in any current or future media, including reprinting/republishing this material for advertising or promotional purposes, creating new collective works, for resale or redistribution to servers or lists, or reuse of any copyrighted component of this work in other works.This manuscript has been submitted to an IEEE Transactions journal and is currently under review.}
    \fancyhead[R]{}
}
\begin{document}

\title{
BLADE: Relia\underline{B}le Dynamic Hardware-Aware SNN--ANN Boundary
Se\underline{L}ection for Event-B\underline{A}se\underline{D}
Object D\underline{E}tection
}

\author{Mahdi~Taheri, Alwin Paul%
\thanks{Mahdi Taheri  is with Humboldt University of Berlin, Germany and also with Tallinn University of Technology, Tallinn, Estonia. Alwin Paul is with Brandenburgische Technische Universität Cottbus-Senftenberg, Germany. (e-mail: mahdi.taheri@Taltech.ee; paulalwi@b-tu.de).}}


\maketitle
\thispagestyle{firstpage}

\begin{abstract}
Hybrid Spiking Neural Network (SNN)--Artificial Neural Network (ANN) architectures combine the energy efficiency of SNNs with the superior detection accuracy of ANNs for event-based object detection. Existing hybrid SNN--ANN networks, however, employ static inference and select the SNN--ANN boundary primarily according to accuracy and energy consumption, without considering dynamic inference or reliability. This paper presents \textsc{BLADE}, the first reliability-aware boundary selection methodology for dynamic hybrid SNN--ANN networks with ANN early exit. The proposed framework jointly optimizes the SNN--ANN boundary and ANN early-exit configuration according to reliability, detection accuracy, execution time, and energy consumption, while incorporating reliability through hierarchical statistical fault injection during design-space exploration. Experimental evaluation on an event-based object detector achieves an mAP@0.5 of $0.691$ while reducing the inference compute energy to $15.82$~mJ when the ANN early exit fires. Reliability analysis identifies the most significant floating-point exponent bit as the dominant source of catastrophic failures, producing significant-or-worse accuracy degradation in $58.8\%$ of its fault injections. Protecting this single bit with approximately $3\%$ storage overhead eliminates catastrophic failures across the evaluated realistic technology fault rates. Furthermore, increasing the proportion of SNN computation improves fault tolerance, with the fully SNN configuration achieving a reliability retention of $0.965$ under aggressive fault conditions. The results demonstrate that jointly optimizing reliability, accuracy, execution time, and energy consumption enables more dependable deployment of dynamic hybrid SNN--ANN systems for safety-critical edge AI applications.
\end{abstract}


\section{Introduction}
\label{sec:introduction}

Object detection is a key workload in edge AI perception, where modern architectures achieve strong performance on automotive and event-based vision benchmarks~\cite{Gehrig_2023_CVPR,yang2025smamba,silva2025chimera}. Practical deployment, however, requires strict execution-time, energy, and reliability constraints, particularly in safety-critical applications.

Spiking Neural Networks (SNNs) reduce computational complexity and energy consumption on neuromorphic hardware through event-driven sparse computation~\cite{sekonji2026fpga,sharifian2025reliability}. However, fully spiking networks still struggle to match the accuracy of conventional Artificial Neural Networks (ANNs). Hybrid SNN--ANN networks address this tradeoff by combining SNN-based early feature extraction with ANN-based detection stages, as in the attention-based SNN--ANN bridge (ASAB)~\cite{ahmed2025efficient} and earlier fixed-boundary hybrids~\cite{kugele2021hybrid}. More broadly, spiking and hybrid neuromorphic systems have demonstrated real-world perception and control on dedicated hardware~\cite{pei2019towards,paredes2024fully}.

The transition boundary between SNN and ANN components strongly affects detection accuracy, execution time, energy consumption, and the workload mapped to each hardware substrate. Existing methods rely on manual boundary selection, limited exploration, or retraining-intensive optimization. For example,~\cite{ahmed2025efficient} evaluates only three partitions of the same backbone, while differentiable search methods such as SpikeDHS~\cite{che2022differentiable} and its spatial-temporal extension~\cite{che2024spatial} do not consider contiguous hardware partitioning. Other studies target different domains~\cite{negi2024best} or shallow architectures requiring retraining for each candidate~\cite{seekings2024hybrid}. Thus, a lightweight post-training methodology for SNN--ANN boundary selection remains unavailable.

Dynamic inference further reduces computation by adapting the executed workload to input complexity~\cite{han2021dynamic}. Early-exit mechanisms terminate inference once sufficient confidence is reached~\cite{teerapittayanon2016branchynet}, while SNN-specific methods dynamically adjust simulation timesteps or terminate spike propagation~\cite{li2023seenn,wu2023topkcutoff}. However, dynamic early exit remains unexplored for hybrid SNN--ANN networks, where both the SNN--ANN boundary and executed depth are typically fixed for all inputs.

Reliability is also critical in the target deployment scenarios of hybrid SNN--ANN networks. Hardware-induced soft errors caused by radiation, process variations, and technology scaling can corrupt network parameters during inference, causing silent accuracy degradation or catastrophic prediction failures~\cite{li2017understanding,riera2016ber}. Since SNN and ANN components use different numerical representations and hardware substrates, different boundary locations expose different fault characteristics. Existing boundary selection methods optimize accuracy and efficiency, but largely ignore reliability.

This paper presents \textsc{BLADE} (Relia\underline{B}le Dynamic Hardware-Aware SNN--ANN Boundary Se\underline{L}ection for Event-B\underline{A}se\underline{D} Object D\underline{E}tection), the first reliability-aware boundary selection methodology for dynamic hybrid SNN--ANN networks with early exit. \textsc{BLADE} jointly optimizes the SNN--ANN boundary and ANN early-exit configuration according to reliability, accuracy, execution time, and energy consumption, while applying bit-level protection as a fixed selective intervention to every candidate. Reliability is quantified through hierarchical statistical fault injection and integrated into the boundary selection process.

The main contributions are summarized as follows:

\begin{itemize}

\item A dynamic hybrid SNN--ANN object detection architecture with early exit enables adaptive inference according to input complexity.

\item A reliability-aware boundary selection methodology jointly optimizes the SNN--ANN boundary and ANN early-exit configuration with respect to reliability, accuracy, execution time, and energy consumption, while applying fixed bit-level protection to every candidate.

\item A hierarchical statistical fault-injection methodology characterizes reliability and identifies dominant fault mechanisms across hybrid SNN--ANN components.

\item Validation on an event-based hybrid object detector shows that reliability-aware boundary selection improves deployment dependability while maintaining competitive detection accuracy and energy efficiency.

\end{itemize}

The remainder of this paper is organized as follows. Section II reviews related work on hybrid SNN--ANN networks, dynamic inference, and reliability assessment. Section III presents the proposed methodology. Section IV reports the experimental results. Section V concludes the paper.

\section{Related Work}
\label{sec:related_work}

Hybrid SNN--ANN networks have emerged as an effective solution for event-based perception by combining the energy efficiency of Spiking Neural Networks (SNNs) with the superior detection accuracy of Artificial Neural Networks (ANNs). Recent event-based object detectors employ an SNN backbone together with an ANN detection network connected through an SNN--ANN bridge~\cite{ahmed2025efficient,pei2019towards,paredes2024fully}. In these architectures, the SNN--ANN boundary determines the computational workload assigned to each substrate and therefore directly affects detection accuracy, execution time, and energy consumption. Existing approaches, however, rely on manually selected boundaries or limited design exploration, with the boundary optimized primarily according to accuracy and efficiency~\cite{ahmed2025efficient,che2022differentiable,che2024spatial}.

Dynamic inference has demonstrated that many inputs do not require the full computational capacity of a neural network~\cite{han2021dynamic}. Early-exit mechanisms and dynamic routing reduce computation by terminating inference once sufficient prediction confidence is reached or by skipping unnecessary network blocks~\cite{teerapittayanon2016branchynet,wu2018blockdrop,wang2018skipnet}. Similar concepts have also been explored for SNNs by reducing simulation timesteps or terminating spike propagation early~\cite{li2023seenn,wu2023topkcutoff}. However, these approaches operate within a single computational substrate and do not address dynamic inference in hybrid SNN--ANN networks.
Reliability assessment of neural networks is commonly performed through statistical fault injection~\cite{leveugle2009,wilson1927,reagen2018ares,ruospo2025iterative}. Previous studies have investigated the vulnerability of floating-point representations~\cite{li2017understanding,msetcep2026}, the reliability of SNNs~\cite{spyrou2024spikefi,putra2022softsnn}, and reliability-aware design-space exploration for deep neural network (DNN) accelerators~\cite{deepaxe2023,saffira2024}. Nevertheless, these studies focus on neural network models or hardware accelerators independently and do not consider the SNN--ANN boundary or ANN early-exit configuration as reliability-aware optimization variables.

Overall, existing research treats hybrid SNN--ANN design, dynamic inference, and reliability assessment as separate problems. Existing hybrid networks employ static inference and optimize the SNN--ANN boundary mainly according to accuracy and energy, while reliability studies characterize fault behavior without incorporating it into deployment decisions. \textsc{BLADE} addresses this gap by jointly optimizing the SNN--ANN boundary and ANN early-exit configuration according to reliability, detection accuracy, execution time, and energy consumption, enabling reliability-aware deployment of dynamic hybrid SNN--ANN networks.

\section{Proposed Methodology}
\label{sec:methodology}
\definecolor{snngreen}{RGB}{39,128,73}
\definecolor{asabred}{RGB}{200,55,45}
\definecolor{headviolet}{RGB}{120,70,160}
\definecolor{busblue}{RGB}{45,95,180}
\definecolor{eeorange}{RGB}{217,108,28}
\definecolor{relred}{RGB}{200,55,55}
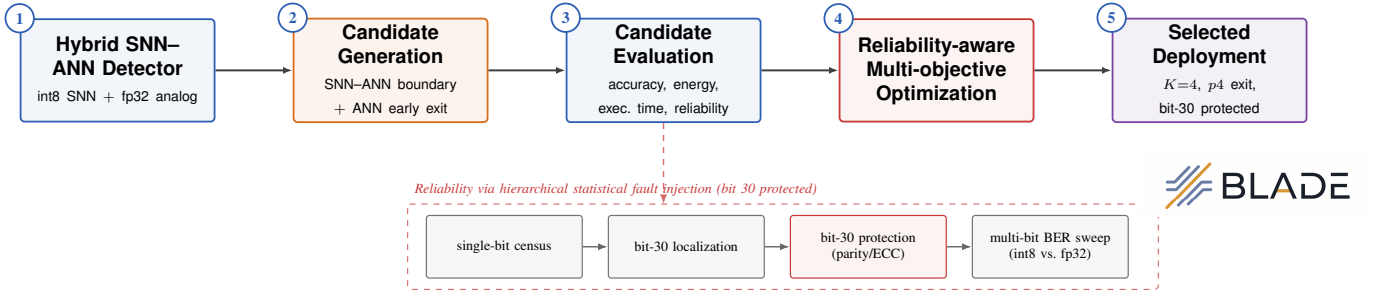
\begin{figure*}[t]
\centering
\resizebox{\textwidth}{!}{%
\begin{tikzpicture}[
  font=\sffamily,
  >={Latex[length=1.7mm]},
  mbox/.style={rounded corners=2pt,draw=busblue,fill=busblue!6,line width=0.9pt,
               minimum height=18mm,text width=32mm,align=center,inner sep=3pt},
  msub/.style={rounded corners=1.5pt,draw=black!55,fill=black!3,line width=0.6pt,
               minimum height=11mm,text width=26mm,align=center,inner sep=2pt,
               font=\scriptsize},
  mbadge/.style={circle,draw=busblue,fill=white,line width=0.9pt,inner sep=0.5pt,
                 font=\footnotesize\bfseries,text=busblue,minimum size=5mm},
  mflow/.style={->,line width=1pt,draw=black!75},
  msflow/.style={->,line width=0.7pt,draw=black!60},
]
\node[mbox] (s1) at (0,0)
  {\textbf{Hybrid SNN--ANN Detector}\\[1pt]{\scriptsize int8 SNN $+$ fp32 analog}};
\node[mbox,draw=eeorange,fill=eeorange!8] (s2) at (4.8,0)
  {\textbf{Candidate Generation}\\[1pt]{\scriptsize SNN--ANN boundary $+$ ANN early exit}};
\node[mbox] (s3) at (9.6,0)
  {\textbf{Candidate Evaluation}\\[1pt]{\scriptsize accuracy, energy, exec.\ time, reliability}};
\node[mbox,draw=relred,fill=relred!6] (s4) at (14.4,0)
  {\textbf{Reliability-aware}\\\textbf{Multi-objective Optimization}};
\node[mbox,draw=headviolet,fill=headviolet!8] (s5) at (19.2,0)
  {\textbf{Selected Deployment}\\[1pt]{\scriptsize $K{=}4$, $p4$ exit, bit-30 protected}};

\foreach \a/\b in {s1/s2,s2/s3,s3/s4,s4/s5}{\draw[mflow] (\a.east) -- (\b.west);}
\foreach \n/\s in {1/s1,2/s2,3/s3,4/s4,5/s5}{\node[mbadge] at (\s.north west) {\n};}

\node[msub] (c1) at (6.8,-3.1) {single-bit census};
\node[msub] (c2) at (10.0,-3.1) {bit-30 localization};
\node[msub,draw=relred,fill=relred!6] (c3) at (13.2,-3.1) {bit-30 protection (parity/ECC)};
\node[msub] (c4) at (16.4,-3.1) {multi-bit BER sweep (int8 vs.\ fp32)};
\foreach \a/\b in {c1/c2,c2/c3,c3/c4}{\draw[msflow] (\a.east) -- (\b.west);}
\draw[draw=relred!70,dashed,rounded corners=2pt,line width=0.6pt]
  (5.1,-2.35) rectangle (18.3,-3.85);
\node[font=\scriptsize\itshape,text=relred,anchor=south west] at (5.1,-2.33)
  {Reliability via hierarchical statistical fault injection (bit~30 protected)};
\draw[msflow,relred!75,dashed] (s3.south) -- (9.6,-2.35);
\node (logo) at (20,-2)
    {\includegraphics[width=4.0cm]{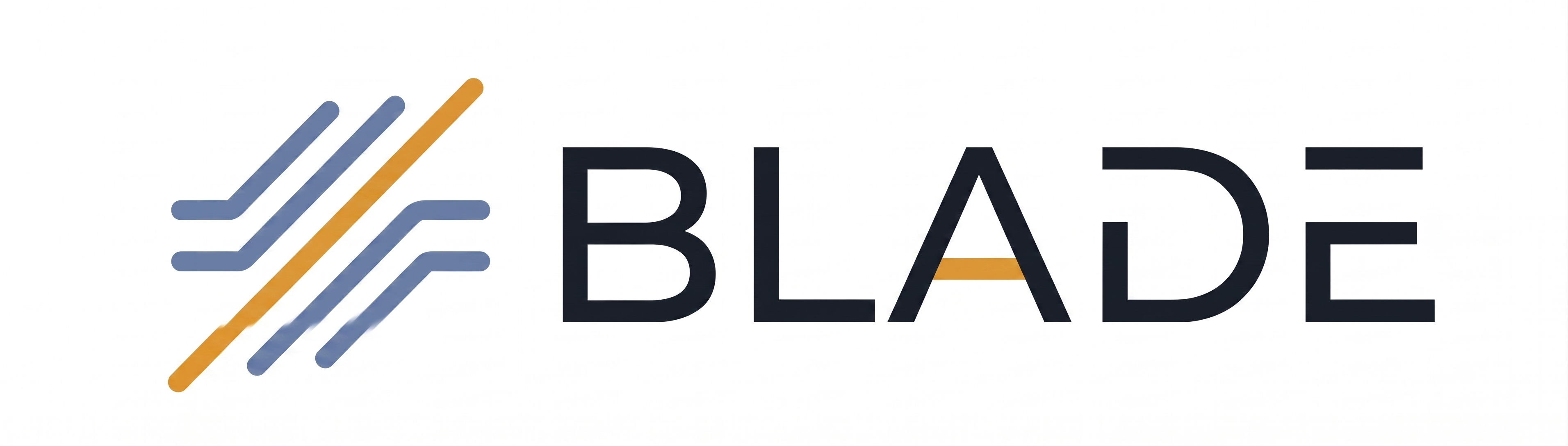}};
\end{tikzpicture}}
\caption{Overview of the proposed BLADE framework: candidate SNN--ANN boundary and ANN early-exit configurations are generated, evaluated on accuracy, energy, execution time, and reliability, and selected by multi-objective optimization.}
\label{fig:methodology}
\end{figure*}

\definecolor{snngreen}{RGB}{39,128,73}
\definecolor{asabred}{RGB}{200,55,45}
\definecolor{headviolet}{RGB}{120,70,160}
\definecolor{busblue}{RGB}{45,95,180}
\definecolor{eeorange}{RGB}{217,108,28}
\definecolor{relred}{RGB}{200,55,55}
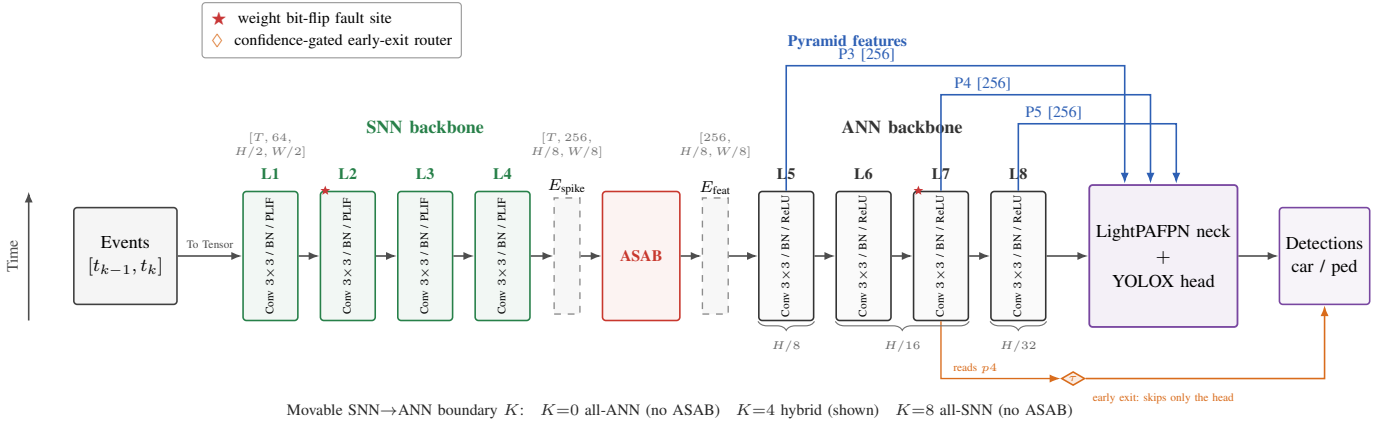
\begin{figure*}[t]
\centering
\resizebox{\textwidth}{!}{%
\begin{tikzpicture}[
  font=\sffamily,
  >={Latex[length=1.7mm]},
  block/.style={rounded corners=1.5pt,line width=0.7pt,minimum height=20mm,
                minimum width=8.5mm,inner sep=1pt},
  snn/.style={block,draw=snngreen,fill=snngreen!6},
  ann/.style={block,draw=black!80,fill=black!2},
  io/.style={rounded corners=1.5pt,draw=black!75,fill=black!4,line width=0.7pt,
             minimum height=15mm,align=center,inner sep=2pt},
  head/.style={rounded corners=1.5pt,draw=headviolet,fill=headviolet!8,
               line width=0.8pt,minimum height=22mm,align=center,inner sep=3pt},
  feat/.style={draw=black!45,dashed,fill=black!3,line width=0.6pt,
               minimum height=18mm,minimum width=4mm,inner sep=0pt},
  op/.style={rotate=90,font=\tiny,inner sep=0pt},
  stg/.style={font=\scriptsize\bfseries,inner sep=1pt},
  dim/.style={font=\tiny,text=black!55,align=center},
  flow/.style={->,line width=0.7pt,draw=black!70},
]

\node[io,minimum width=16mm,font=\footnotesize] (ev) at (1.7,0) {Events\\[1pt]$[t_{k-1},t_k]$};

\foreach \i/\x in {1/3.95,2/5.15,3/6.35,4/7.55}{
  \node[snn] (L\i) at (\x,0) {};
  \node[op] at (\x,0) {Conv $3{\times}3$ / BN / PLIF};
  \node[stg,text=snngreen] at (\x,1.28) {L\i};
}
\node[feat] (esp) at (8.55,0) {};
\node[block,draw=asabred,fill=asabred!6,minimum width=12mm] (asab) at (9.7,0) {};
\node[font=\scriptsize\bfseries,text=asabred,align=center] at (9.7,0) {ASAB};
\node[feat] (efe) at (10.85,0) {};

\foreach \i/\x in {5/11.95,6/13.15,7/14.35,8/15.55}{
  \node[ann] (L\i) at (\x,0) {};
  \node[op] at (\x,0) {Conv $3{\times}3$ / BN / ReLU};
  \node[stg,text=black!80] at (\x,1.28) {L\i};
}
\node[head,minimum width=16mm,font=\footnotesize] (head) at (17.8,0) {LightPAFPN neck\\[2pt]$+$\\[2pt]YOLOX head};
\node[io,minimum width=14mm,draw=headviolet,fill=headviolet!6,font=\footnotesize] (det) at (20.3,0)
  {Detections\\[1pt]car / ped};

\draw[flow] (ev) -- node[above,font=\tiny,text=black!70]{To Tensor} (L1);
\draw[flow] (L1)--(L2); \draw[flow] (L2)--(L3); \draw[flow] (L3)--(L4);
\draw[flow] (L4)--(esp); \draw[flow] (esp)--(asab); \draw[flow] (asab)--(efe);
\draw[flow] (efe)--(L5);
\draw[flow] (L5)--(L6); \draw[flow] (L6)--(L7); \draw[flow] (L7)--(L8);
\draw[flow] (L8.east) -- (head.west);
\draw[flow] (head)--(det);
\node[draw=eeorange,diamond,aspect=1.7,inner sep=0.5pt,line width=0.7pt,
      fill=eeorange!12,font=\tiny,text=eeorange] (gate) at (16.4,-1.9) {$\tau$};
\draw[eeorange,->,line width=0.6pt] (L7.south) |- (gate.west);
\node[font=\tiny,text=eeorange,inner sep=1pt,anchor=south west] at (14.5,-1.86)
  {reads $p4$};
\draw[eeorange,->,line width=0.7pt] (gate.east) -| (det.south);
\node[font=\tiny,text=eeorange,inner sep=1pt,anchor=north] at (17.8,-2.08)
  {early exit: skips only the head};

\node[font=\scriptsize,above=0pt of esp,inner sep=1pt] {$E_{\text{spike}}$};
\node[font=\scriptsize,above=0pt of efe,inner sep=1pt] {$E_{\text{feat}}$};

\node[font=\footnotesize\bfseries,text=snngreen] at (6.35,2.0) {SNN backbone};
\node[font=\footnotesize\bfseries,text=black!80] at (13.75,2.0) {ANN backbone};
\node[dim] at (3.95,1.72) {$[T,64,$\\$H/2,W/2]$};
\node[dim] at (8.55,1.72) {$[T,256,$\\$H/8,W/8]$};
\node[dim] at (10.85,1.72) {$[256,$\\$H/8,W/8]$};

\draw[decorate,decoration={brace,amplitude=4pt,mirror},draw=black!55,line width=0.5pt]
  (L5.south west) -- (L5.south east) node[midway,below=5pt,font=\tiny,text=black!60]{$H/8$};
\draw[decorate,decoration={brace,amplitude=4pt,mirror},draw=black!55,line width=0.5pt]
  (L6.south west) -- (L7.south east) node[midway,below=5pt,font=\tiny,text=black!60]{$H/16$};
\draw[decorate,decoration={brace,amplitude=4pt,mirror},draw=black!55,line width=0.5pt]
  (L8.south west) -- (L8.south east) node[midway,below=5pt,font=\tiny,text=black!60]{$H/32$};

\draw[busblue,line width=0.7pt,->] (L5.north) |- (17.2,2.95) -- (17.2,1.12);
\draw[busblue,line width=0.7pt,->] (L7.north) |- (17.6,2.50) -- (17.6,1.12);
\draw[busblue,line width=0.7pt,->] (L8.north) |- (18.0,2.05) -- (18.0,1.12);
\node[busblue,font=\scriptsize\bfseries] at (12.9,3.32) {Pyramid features};
\node[busblue,font=\scriptsize] at (13.2,3.12) {P3 [256]};
\node[busblue,font=\scriptsize] at (15.0,2.67) {P4 [256]};
\node[busblue,font=\scriptsize] at (16.1,2.22) {P5 [256]};

\node[star,star points=5,star point ratio=2.3,fill=relred,inner sep=0pt,
      minimum size=4.5pt] (snnfi) at ([xshift=-3.5mm]L2.north) {};
\node[star,star points=5,star point ratio=2.3,fill=relred,inner sep=0pt,
      minimum size=4.5pt] (annfi) at ([xshift=-3.5mm]L7.north) {};
\node[draw=black!40,rounded corners=1.5pt,line width=0.5pt,inner sep=4pt,
      align=left,font=\scriptsize] at (4.9,3.5)
  {\textcolor{relred}{$\bigstar$}~~weight bit-flip fault site\\[2pt]
   \textcolor{eeorange}{$\Diamond$}~~confidence-gated early-exit router};


\draw[->,line width=0.7pt,draw=black!70] (0.2,-1.0) -- (0.2,1.0);
\node[rotate=90,font=\scriptsize,text=black!70] at (-0.05,0) {Time};

\node[font=\scriptsize,text=black!85,align=center] at (10.3,-2.4)
  {Movable SNN$\rightarrow$ANN boundary $K$:\quad
   $K{=}0$ all-ANN (no ASAB)\quad $K{=}4$ hybrid (shown)\quad $K{=}8$ all-SNN (no ASAB)};

\end{tikzpicture}}
\caption{Dynamic early-exit hybrid SNN--ANN event-camera detector ($K{=}4$
shown) with movable SNN$\rightarrow$ANN boundary $K$; weight bit-flip
fault-injection sites and the confidence-gated early-exit router are marked
(\Cref{fig:early_exit}; key at upper left).}
\label{fig:architecture}
\end{figure*}
 
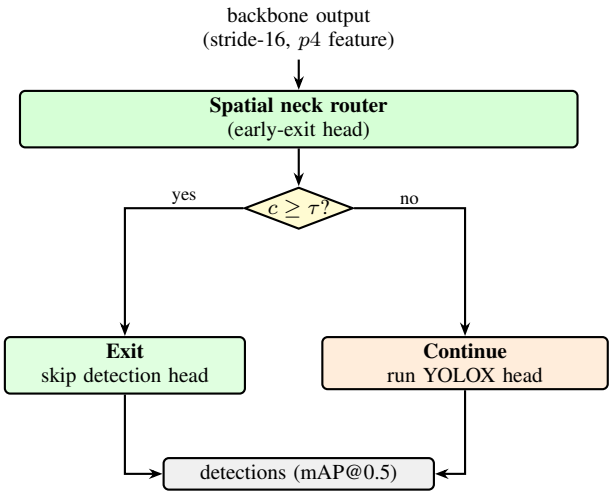
\begin{figure}[t]
\centering
\begin{tikzpicture}[
  font=\footnotesize,
  >={Stealth[length=1.8mm]},
  box/.style={draw, rounded corners=2pt, thick, align=center, inner sep=2.5pt},
  router/.style={box, fill=green!16, text width=7.2cm},
  gate/.style={draw, thick, diamond, aspect=2.6, align=center, fill=yellow!22, inner sep=0pt},
  exitb/.style={box, fill=green!12, text width=3.0cm},
  contb/.style={box, fill=orange!14, text width=3.6cm},
  outb/.style={box, fill=gray!12, text width=3.4cm},
  arr/.style={->, thick},
  lbl/.style={font=\scriptsize, inner sep=1pt},
]
\node[align=center] (in) {backbone output\\(stride-16, $p4$ feature)};
\node[router, below=4mm of in] (router)
  {\textbf{Spatial neck router}\\(early-exit head)};
\node[gate, below=5mm of router] (gate) {$c \ge \tau$?};
\node[exitb, below=14mm of gate, xshift=-23mm] (exit)
  {\textbf{Exit}\\skip detection head};
\node[contb, below=14mm of gate, xshift=22mm] (cont)
  {\textbf{Continue}\\run YOLOX head};
\node[outb, below=30mm of gate] (out) {detections (mAP@0.5)};

\draw[arr] (in) -- (router);
\draw[arr] (router) -- (gate);
\draw[arr] (gate.west) -| (exit.north) node[lbl, pos=0.25, above]{yes};
\draw[arr] (gate.east) -| (cont.north) node[lbl, pos=0.25, above]{no};
\draw[arr] (exit.south) |- (out.west);
\draw[arr] (cont.south) |- (out.east);
\end{tikzpicture}
\caption{Dynamic ANN early-exit mechanism: the spatial neck router exits at the stride-16 ($p4$) feature. The diamond shows the $c\ge\tau$ confidence baseline; the deployed exit is taken by the learned controller (Results).}
\label{fig:early_exit}
\end{figure}
  
\begin{figure}[t]
\centering
\begin{tikzpicture}[
  font=\sffamily\scriptsize,
  >={Stealth[length=1.6mm]},
  cell/.style={draw, minimum width=3.6mm, minimum height=5.2mm, inner sep=0pt, anchor=south west},
  sign/.style={cell, fill=blue!16},
  expo/.style={cell, fill=cyan!22},
  msb/.style={cell, fill=red!75, text=white, font=\sffamily\scriptsize\bfseries},
  frac/.style={cell, fill=black!8},
  fld/.style={font=\sffamily\tiny, inner sep=1pt},
  capt/.style={font=\sffamily\scriptsize\bfseries, inner sep=1pt},
]
\def\w{3.6mm}
\node[capt, anchor=south west] at (1.7*\w,1.16) {fp32 analog weight (ANN, ASAB, neck, head)};
\node[sign] (s) at (0,0) {};
\node[msb]  (e30) at (1*\w,0) {30};
\foreach \i/\b in {2/29,3/28,4/27,5/26,6/25,7/24,8/23}
  \node[expo] (e\b) at (\i*\w,0) {};
\node[frac, minimum width=9*\w, anchor=south west] (f) at (9*\w,0) {};
\node[fld, anchor=north] at ($(s.south)+(0,-0.2mm)$) {31};
\node[fld, anchor=south] at ($(s.north)+(0,0.4mm)$) {S};
\draw[decorate, decoration={brace, amplitude=1.0mm, mirror}]
  ([yshift=-1.0mm]e30.south west) -- node[below=0.8mm, fld]{exponent: bits 30--23} ([yshift=-1.0mm]e23.south east);
\draw[decorate, decoration={brace, amplitude=1.0mm, mirror}]
  ([yshift=-1.0mm]f.south west) -- node[below=0.8mm, fld]{mantissa: bits 22--0} ([yshift=-1.0mm]f.south east);

\begin{scope}[yshift=-21mm]
\node[capt, anchor=south west] at (1.7*\w,1.16) {int8 spiking weight (SNN backbone, Loihi~2)};
\node[sign] (is) at (0,0) {};
\foreach \i/\b in {1/6,2/5,3/4,4/3,5/2,6/1,7/0}
  \node[expo] (ib\b) at (\i*\w,0) {};
\node[fld, anchor=south] at ($(is.north)+(0,0.4mm)$) {b7};
\draw[decorate, decoration={brace, amplitude=1.0mm, mirror}]
  ([yshift=-1.0mm]is.south west) -- node[below=0.8mm, fld]{8-bit signed integer (no exponent field)} ([yshift=-1.0mm]ib0.south east);
\end{scope}

\begin{scope}[yshift=-31.5mm]
\node[sign, minimum width=3.0mm, minimum height=3.0mm] (l1) at (0,0) {};
\node[fld, anchor=west, font=\sffamily\scriptsize] at (l1.east) {sign};
\node[expo, minimum width=3.0mm, minimum height=3.0mm] (l2) at (1.25,0) {};
\node[fld, anchor=west, font=\sffamily\scriptsize] at (l2.east) {exponent (fp32) / integer (int8)};
\node[msb, minimum width=3.0mm, minimum height=3.0mm, font=\sffamily] (l3) at (0,-0.5) {};
\node[fld, anchor=west, font=\sffamily\scriptsize] at (l3.east) {exponent MSB (protected)};
\node[frac, minimum width=3.0mm, minimum height=3.0mm] (l4) at (3.9,-0.5) {};
\node[fld, anchor=west, font=\sffamily\scriptsize] at (l4.east) {mantissa};
\end{scope}
\end{tikzpicture}
\caption{Weight bit fields of the fp32 analog and int8 spiking substrates: the
fp32 exponent MSB (bit~30) is the single protectable catastrophic site. Per-field
single-bit census in \Cref{tab:bitfield}.}
\label{fig:float_protection}
\end{figure}

This section presents the proposed BLADE framework for reliability-aware boundary selection in dynamic hybrid SNN--ANN object detectors. The framework jointly explores the design space defined by the SNN--ANN boundary and the ANN early-exit configuration to identify deployment solutions that satisfy different design objectives. Unlike conventional boundary-selection methods that primarily consider accuracy and energy consumption and were manual, partial, or retraining-heavy, the proposed methodology is fully automated and also incorporates execution time and reliability into the optimization process. Reliability is quantified through hierarchical statistical fault injection and directly integrated into the boundary selection procedure.

The complete workflow of BLADE is illustrated in Fig.~\ref{fig:methodology}.
Starting from a trained hybrid SNN--ANN detector, the framework first constructs all valid combinations of SNN--ANN boundaries and ANN early-exit configurations. Each candidate is then evaluated according to four complementary metrics, namely accuracy, energy consumption, execution time, and reliability. The resulting measurements are combined through a multi-objective optimization process that identifies the most suitable deployment configuration according to the selected design priorities.

\subsection{Hybrid SNN--ANN Search Space}
\label{sec:searchspace}

The proposed methodology considers a hybrid object detector composed of an SNN backbone followed by an ANN detection network. The SNN processes the event stream and extracts temporal features, while the ANN performs feature refinement and object detection. The transition between the two computational substrates is controlled by the boundary variable $K$, which determines the number of backbone stages executed as SNN layers.

Let the backbone consist of eight sequential stages. A boundary configuration is denoted by

\[
S_{K}A_{8-K},
\]

where the first $K$ stages execute as SNN blocks and the remaining $8-K$ stages execute as ANN blocks. The search space considered in this work contains

\[
K \in \{0,2,4,8\},
\]

where

\begin{itemize}
\item $K=0$ corresponds to a fully ANN backbone,
\item $K=8$ corresponds to a fully SNN backbone,
\item intermediate values represent hybrid SNN--ANN realizations.
\end{itemize}

The candidate set covers both substrate endpoints together with two intermediate hybrids, which is sufficient to expose how the boundary location trades detection accuracy, energy consumption, and reliability across the full SNN--ANN continuum. The set is kept compact because each candidate boundary requires an independently trained detector together with its own statistical fault-injection and energy characterization, so enumerating every integer boundary would multiply the characterization cost without changing the observed monotonic trends.

This contiguous partition reflects the hardware organization of heterogeneous neuromorphic systems, where a single interface connects the neuromorphic processor and the conventional AI accelerator. Unlike arbitrary per-layer assignments, the contiguous boundary preserves feature compatibility while minimizing communication overhead.
Moreover, the SNN backbone consumes the natively spike-encoded event stream over the $T$ temporal bins, and the ASAB bridge performs a one-way spiking-to-dense conversion; the analog stages therefore operate on the resulting dense feature map and cannot precede the spiking ones, so the spiking stages necessarily form the prefix of the partition.

Let $h_t^{(s)}$ denote the hidden state of spiking stage $s$ at timestep $t$. For an event sequence consisting of $T=10$ temporal bins, the SNN backbone evolves according to

\begin{equation}
h_t^{(s)} = f_s\left(h_{t-1}^{(s)},\,h_t^{(s-1)},\,x_t\right),
\label{eq:plif}
\end{equation}

where $f_s(\cdot)$ represents the parametric leaky integrate-and-fire (PLIF) neuron dynamics and $x_t$ denotes the input event tensor at timestep $t$. After the final SNN stage defined by the selected boundary $K$, the accumulated spiking representation is converted by the ASAB bridge into continuous ANN feature maps that are processed by the remaining ANN stages.

Unlike existing hybrid SNN--ANN detectors, the proposed architecture (Fig.~\ref{fig:architecture}) further introduces dynamic inference through ANN early exits. The SNN backbone is executed completely for every input sample to preserve temporal feature extraction. Dynamic execution is introduced only after the SNN--ANN boundary, where multiple candidate exit locations are inserted within the ANN component. Depending on the confidence of the intermediate prediction, inference either terminates at the selected ANN exit or continues toward the final detection head. Consequently, each deployment candidate is uniquely defined by two design variables:

\[
\mathcal{C} = (K,E),
\]

where $K$ denotes the SNN--ANN boundary and $E$ denotes the selected ANN early-exit location.

The complete design space is therefore generated by enumerating all valid combinations of boundary positions and ANN early-exit configurations. Each candidate is subsequently evaluated according to accuracy, energy consumption, execution time, and reliability before the final deployment decision is made.

\subsection{Candidate Evaluation}
\label{sec:candidate_evaluation}

Each candidate configuration $\mathcal{C}=(K,E)$ generated in the previous subsection is evaluated according to four complementary metrics, namely energy consumption, execution time, detection accuracy, and reliability. Unlike existing boundary-selection approaches that optimize only a subset of these objectives, the proposed methodology quantifies all four metrics for every candidate before the final optimization stage. Since the SNN--ANN boundary and the ANN early-exit location jointly determine the executed computation graph, every candidate exhibits a unique energy consumption, execution time, accuracy, and reliability profile.

\subsubsection{Energy Model}
\label{sec:energy_model}

The energy consumption of each candidate is estimated using a heterogeneous energy model that separately accounts for the computations executed on the neuromorphic processor and the conventional ANN accelerator. Since the SNN and ANN parts employ different computational primitives, their energy costs are modeled independently.

For each backbone layer $\ell$, the execution energy is

\begin{equation}
E_\ell =
\begin{cases}
r_{\ell}^{\mathrm{in}} \cdot \mathrm{MAC}_\ell \cdot T \cdot c_{\mathrm{AC}}
+
n_\ell \cdot T \cdot c_n,
&
\ell \in \mathrm{SNN}
\\
\mathrm{MAC}_\ell \cdot c_{\mathrm{MAC}},
&
\ell \in \mathrm{ANN},
\end{cases}
\label{eq:layer_energy}
\end{equation}

where $\mathrm{MAC}_\ell$ denotes the number of multiply-accumulate operations of layer $\ell$, $r_{\ell}^{\mathrm{in}}$ is the average input spike activity, $T$ is the number of simulation timesteps, $n_\ell$ is the number of neurons, $c_{\mathrm{AC}}$ is the energy per accumulate operation on the neuromorphic processor, $c_n$ is the neuron-update energy, and $c_{\mathrm{MAC}}$ is the energy per multiply-accumulate operation on the conventional accelerator.

The ASAB bridge is always executed on the ANN processor and therefore contributes

\begin{equation}
E_{\mathrm{bridge}}
=
\mathrm{MAC}_{\mathrm{ASAB}}
\cdot
c_{\mathrm{MAC}}.
\end{equation}

The detection head is likewise executed entirely on the ANN processor and contributes a constant energy independent of the selected boundary. Consequently, the total energy of candidate $(K,E)$ is

\begin{equation}
E_{\mathrm{tot}}(K,E)
=
\sum_{\ell=1}^{K}
E_\ell^{\mathrm{SNN}}
+
\sum_{\ell=K+1}^{E}
E_\ell^{\mathrm{ANN}}
+
E_{\mathrm{bridge}}
+
E_{\mathrm{exit}}(E),
\label{eq:energy_total}
\end{equation}

where $E_{\mathrm{exit}}(E)$ denotes the computational cost associated with the selected ANN exit.

The proposed model captures arithmetic operations performed by both computational substrates. Memory accesses, static power, and communication overhead are excluded, following the assumptions adopted in previous heterogeneous energy models~\cite{dampfhoffer2022snns,yan2024reconsidering,shen2024conventional}. Since these components remain identical across all candidate configurations, they do not affect the relative ranking performed by BLADE.
The reported energy values are therefore compute-only estimates intended for comparing candidate configurations on a common basis rather than as absolute system-level power figures; translating them to a deployed system would additionally require accounting for memory-access, static-power, and inter-substrate communication costs, which is left to platform-specific characterization.

The heterogeneous model also provides a hardware-dependent crossover condition that determines whether a layer is more efficiently executed as an SNN or ANN layer. Equating the two execution costs in (\ref{eq:layer_energy}) yields

\begin{equation}
r_{\ell}^{\mathrm{in}}
<
\frac{c_{\mathrm{MAC}}}
{T c_{\mathrm{AC}}}
-
\frac{n_\ell c_n}
{\mathrm{MAC}_\ell c_{\mathrm{AC}}},
\label{eq:tau_exact}
\end{equation}

which reduces to

\begin{equation}
r_{\ell}^{\mathrm{in}}
<
\tau(\mathcal{H})
=
\frac{c_{\mathrm{MAC}}}
{T c_{\mathrm{AC}}},
\label{eq:tau}
\end{equation}

when the neuron-update term is negligible. The threshold $\tau(\mathcal{H})$ depends only on the target hardware parameters and therefore indicates the preferred realization of each layer for a given heterogeneous platform.
The energy constants used in this work instantiate one representative heterogeneous accelerator pairing and enter the model only as scalar per-operation costs. Consequently, the boundary-selection procedure is unchanged for alternative pairings, such as other neuromorphic or systolic-array accelerators; a different $c_{\mathrm{AC}}/c_{\mathrm{MAC}}$ ratio simply shifts the crossover $\tau(\mathcal{H})$ and the absolute energy values, while the relative comparison among candidate configurations is expected to transfer.

\subsubsection{ANN Early-Exit Model}
\label{sec:early_exit}

The proposed framework extends conventional hybrid SNN--ANN inference by introducing dynamic execution through ANN early exits (Fig.~\ref{fig:early_exit}). Unlike existing early-exit architectures, the complete SNN backbone is executed for every input sample to preserve temporal feature extraction. Dynamic execution begins only after the SNN--ANN boundary, where multiple candidate exits are inserted within the ANN component.

Each exit consists of an auxiliary detection head attached to an intermediate ANN feature map. During inference, the confidence of the intermediate prediction is evaluated against a predefined threshold $\tau_c$. Samples whose confidence exceeds the threshold terminate inference immediately, while the remaining samples continue toward deeper ANN stages until either another exit is reached or the final detector is executed.
This fixed-threshold rule serves as the baseline operating point. In the deployed framework, the per-sample exit decision is instead produced by a learned controller over the intermediate detection state, with $\tau_c$ parameterizing the corresponding operating point along the accuracy--cost curve.

Let $P_e$ denote the probability that inference terminates at exit $e$. The expected execution energy of candidate $(K,E)$ is therefore

\begin{equation}
\bar{E}(K,E)
=
\mathbb{E}
\left[
E_{\mathrm{tot}}(K,E)
\right]
=
\sum_{e}
P_e
\,
E_e(K),
\label{eq:expected_energy}
\end{equation}

where $E_e(K)$ represents the cumulative energy required to execute the SNN backbone, the ANN layers up to exit $e$, and the corresponding detection head.

Similarly, the expected execution time is

\begin{equation}
\mathbb{E}
\left[
T_{\mathrm{exec}}(K,E)
\right]
=
\sum_{e}
P_e
\,
T_e(K),
\label{eq:expected_latency}
\end{equation}

where $T_e(K)$ denotes the execution latency associated with exit $e$.

Because the SNN backbone is always executed completely, different exit locations only modify the amount of ANN computation performed after the selected SNN--ANN boundary. Consequently, the energy consumption and execution time of each candidate depend jointly on the selected boundary and ANN early-exit location.

The detection accuracy associated with each candidate is obtained directly from inference on the validation dataset. Unlike the energy model, which is estimated analytically, accuracy is measured experimentally and used together with energy, execution time, and reliability during the final multi-objective optimization.

\subsubsection{Reliability Characterization}
\label{sec:reliability_char}

Reliability constitutes the fourth design objective considered by BLADE. Since the selected SNN--ANN boundary determines which parts of the network execute on different computational substrates and numerical representations, each candidate exhibits a different fault behavior. The proposed methodology therefore evaluates the reliability of every candidate configuration through hierarchical statistical fault injection before the final optimization.

The fault-injection campaign is performed directly on the deployed hybrid SNN--ANN network without retraining. Both the SNN and ANN components are evaluated using their native numerical representations, where the SNN backbone employs signed 8-bit integer parameters and neuron states, while the ANN component uses single-precision floating-point parameters. This substrate-resolved evaluation enables the reliability contribution of each computational domain to be quantified independently.

Two complementary fault models are considered. The first model evaluates isolated transient faults through single-bit injections, enabling the vulnerability of individual bit positions, layers, and numerical representations to be identified. The second model evaluates simultaneous hardware faults using technology-dependent Bit Error Rates (BERs), allowing the reliability of the complete network to be assessed under realistic fault conditions.

For the single-bit fault model, one random bit is selected from the target memory element and inverted according to

\begin{equation}
w_f = w \oplus (1 \ll b),
\label{eq:singlebit}
\end{equation}

where $w$ denotes the original parameter value, $b$ represents the selected bit position, and $w_f$ is the corrupted value after injection.

The BER-based model independently corrupts every stored bit with probability

\begin{equation}
P(\mathrm{flip})=\mathrm{BER},
\label{eq:ber}
\end{equation}

allowing multiple simultaneous bit flips to occur during a single experiment. The evaluated BER values correspond to technology-dependent soft-error rates derived from~\cite{riera2016ber}, enabling the reliability of each candidate configuration to be compared under realistic operating conditions.

For every candidate configuration $(K,E)$, the complete fault-injection campaign is repeated over the evaluated bit-error-rate ladder, with multiple random fault realizations drawn at each rate. Let $\mathcal{B}$ denote the set of evaluated BER values, $A_{\mathrm{clean}}(K,E)$ the fault-free detection accuracy, and $A(K,E;\beta)$ the mean detection accuracy obtained under bit-error rate $\beta\in\mathcal{B}$. The reliability score associated with the candidate is computed as

\begin{equation}
R(K,E)
=
\frac{1}{\lvert\mathcal{B}\rvert}
\sum_{\beta\in\mathcal{B}}
\frac{A(K,E;\beta)}
{A_{\mathrm{clean}}(K,E)},
\label{eq:reliability}
\end{equation}

which represents the mean accuracy retention across the evaluated bit-error-rate ladder, relative to the fault-free accuracy.

Since exhaustive fault injection is computationally infeasible, statistical sampling is employed following the methodology proposed by Leveugle~\cite{leveugle2009}. The confidence of the estimated failure probability is quantified using Wilson confidence intervals~\cite{wilson1927}, while the iterative convergence methodology proposed in~\cite{ruospo2025iterative} is adopted to ensure sufficient statistical confidence. This approach provides a practical estimation of reliability while maintaining manageable simulation time.

The proposed hierarchical campaign characterizes reliability at multiple levels, including numerical representation, bit position, network layer, SNN and ANN components, and complete hybrid configurations. Consequently, the reliability score associated with every candidate reflects both the vulnerability of the selected SNN--ANN boundary and the amount of ANN computation executed before the selected early exit.


\subsubsection{Bit-Level Protection}
\label{sec:bit_protection}

The hierarchical reliability characterization identifies the dominant sources of failure within the hybrid SNN--ANN network before the final optimization. Rather than uniformly protecting all stored parameters, the proposed methodology first localizes the most vulnerable bit positions and then evaluates the effect of selective protection.

The single-bit fault campaign reveals that catastrophic failures are highly concentrated within a single floating-point exponent bit of the ANN component, whereas the remaining floating-point bits and the integer representation used by the SNN exhibit substantially lower vulnerability. Consequently, selectively protecting this exponent bit provides a significant improvement in reliability while introducing only a negligible storage overhead.

The protected configuration is incorporated into the candidate evaluation process by repeating the complete fault-injection campaign after excluding the protected bit from the fault space. All candidate configurations are therefore evaluated under identical protection assumptions, allowing the influence of the SNN--ANN boundary and the ANN early-exit configuration to be compared independently of avoidable catastrophic failures.

The resulting protected reliability score is used during the final optimization stage together with the measured detection accuracy, analytical energy consumption, and execution time. Consequently, the selected deployment configuration reflects the intrinsic robustness of each candidate rather than the effect of a single easily protectable hardware vulnerability.

\subsection{Multi-Objective Optimization}
\label{sec:optimization}

After evaluating every candidate configuration, BLADE performs a multi-objective optimization to determine the most suitable deployment configuration according to the target application requirements. Each candidate is characterized by four complementary metrics,

\[
\mathcal{M}(K,E)=
\left\{
A(K,E),
R(K,E),
\bar{E}(K,E),
T_{\mathrm{exec}}(K,E)
\right\},
\]

where $A(K,E)$ denotes the measured detection accuracy, $R(K,E)$ represents the reliability score obtained through statistical fault injection, $\bar{E}(K,E)$ is the expected energy consumption, and $T_{\mathrm{exec}}(K,E)$ denotes the expected execution time.

Unlike conventional boundary-selection techniques that optimize only accuracy or energy, the proposed methodology simultaneously considers all four metrics. Since these objectives exhibit different scales and optimization directions, each metric is first normalized over the complete candidate space.

For metrics that are maximized, namely accuracy and reliability,

\begin{equation}
\hat{A}(K,E)
=
\frac{A(K,E)-A_{\min}}
{A_{\max}-A_{\min}},
\label{eq:acc_norm}
\end{equation}

\begin{equation}
\hat{R}(K,E)
=
\frac{R(K,E)-R_{\min}}
{R_{\max}-R_{\min}},
\label{eq:rel_norm}
\end{equation}

while energy consumption and execution time are normalized as

\begin{equation}
\hat{E}(K,E)
=
1-
\frac{\bar{E}(K,E)-E_{\min}}
{E_{\max}-E_{\min}},
\label{eq:energy_norm}
\end{equation}

\begin{equation}
\hat{T}(K,E)
=
1-
\frac{T_{\mathrm{exec}}(K,E)-T_{\min}}
{T_{\max}-T_{\min}},
\label{eq:time_norm}
\end{equation}

so that larger normalized values always correspond to better candidate quality.

The final objective function is computed as a weighted combination of the normalized metrics,

\begin{equation}
J(K,E)
=
w_A\hat{A}(K,E)
+
w_R\hat{R}(K,E)
+
w_E\hat{E}(K,E)
+
w_T\hat{T}(K,E),
\label{eq:objective}
\end{equation}

subject to

\[
w_A+w_R+w_E+w_T=1.
\]

The weighting coefficients allow the optimization to emphasize different deployment objectives. For example, reliability-critical systems assign a larger value to $w_R$, while energy-constrained applications prioritize $w_E$. Similarly, latency-sensitive applications increase $w_T$, whereas accuracy-oriented deployments assign the largest weight to $w_A$.

The optimal deployment configuration is therefore determined by maximizing the objective over the feasible Pareto set $\mathcal{F}$ (the non-dominated candidates satisfying the accuracy, reliability, and catastrophe-probability constraints),

\begin{equation}
(K^{*},E^{*})
=
\arg\max_{(K,E)\in\mathcal{F}}
J(K,E),
\label{eq:optimal}
\end{equation}

where $(K^{*},E^{*})$ denotes the selected SNN--ANN boundary together with the corresponding ANN early-exit configuration.

Feasibility requires every retained candidate to satisfy $A(K,E)\ge A_{\mathrm{feas}}$, $R(K,E)\ge R_{\mathrm{feas}}$, and $P_{\mathrm{cat}}(K,E)\le P_{\max}$, where the catastrophe probability $P_{\mathrm{cat}}(K,E)$ is the Wilson $95\%$ upper bound on the fraction of fault realizations whose faulted accuracy falls below the catastrophe floor $\mathrm{mAP}<0.10$, and $P_{\max}=0.05$; this catastrophe cap is the constraint that forces the bit-level protection of Section~\ref{sec:bit_protection}.


\subsubsection{Optimization Procedure}

Algorithm~\ref{alg:blade} summarizes the complete BLADE workflow. Starting from a trained hybrid SNN--ANN detector, the framework first generates every valid boundary and ANN early-exit configuration. Each candidate is subsequently evaluated according to the analytical energy model, measured execution time, detection accuracy, and reliability characterization described in the previous subsections. The non-dominated candidates are then retained through a Pareto filter and constrained to those satisfying the accuracy, reliability, and catastrophe-probability requirements, after which the normalized objective score is computed over this feasible set and the configuration with the highest value is selected for deployment.

Unlike exhaustive design-space exploration approaches that require repeated retraining for every candidate, BLADE evaluates all candidate configurations within a unified framework and selects the deployment solution directly from the measured design-space characteristics.


\begin{algorithm}[t]
\caption{BLADE: Reliability-Aware Boundary and ANN Early-Exit Selection}
\label{alg:blade}

\begin{algorithmic}[1]

\Require
Trained hybrid SNN--ANN detector,
candidate boundaries $\mathcal{K}$,
candidate ANN exits $\mathcal{E}$,
objective weights
$\{w_A,w_R,w_E,w_T\}$,
feasibility thresholds
$\{A_{\mathrm{feas}},R_{\mathrm{feas}},P_{\max}\}$

\Ensure
Optimal deployment configuration
$(K^{*},E^{*})$

\ForAll{$K\in\mathcal{K}$}

    \ForAll{$E\in\mathcal{E}$}
        \ForAll{$\mathrm{protect}\in\{\text{off},\text{on}\}$}

            \State Generate candidate $c=(K,E,\mathrm{protect})$

            \State Estimate expected energy $\bar{E}(c)$ and execution time $T_{\mathrm{exec}}(c)$ 

            \State Evaluate detection accuracy $A(c)$

            \State Look up reliability $R(c)$ and catastrophe probability $P_{\mathrm{cat}}(c)$ from the offline fault-injection characterization

        \EndFor
    \EndFor

\EndFor

\State $\mathcal{P}\gets\textsc{ParetoFront}(\{c\})$ \Comment{non-dominated on $(R\!\uparrow,A\!\uparrow,E\!\downarrow,T\!\downarrow)$}

\State $\mathcal{F}\gets\{c\in\mathcal{P} : A(c)\!\ge\!A_{\mathrm{feas}},\ R(c)\!\ge\!R_{\mathrm{feas}},\ P_{\mathrm{cat}}(c)\!\le\!P_{\max}\}$ \Comment{feasible set; $P_{\max}$ forces bit-30 protection}

\State Min--max normalize $\{A,R,\bar{E},T_{\mathrm{exec}}\}$ over $\mathcal{F}$ and compute $J(c)$ for $c\in\mathcal{F}$

\State

\[
(K^{*},E^{*})
=
\arg\max_{c\in\mathcal{F}} J(c)
\]

\State \Return
$(K^{*},E^{*})$

\end{algorithmic}

\vspace{2pt}\hfill\includegraphics[height=1.8em]{BLADE_logo.png}\vspace{-2pt}
\end{algorithm}

\subsection{Deployment Configuration}
\label{sec:deployment}

The final stage of BLADE deploys the candidate configuration that maximizes the multi-objective optimization function. The selected deployment is defined by the optimal SNN--ANN boundary $K^{*}$ together with its corresponding ANN early-exit configuration $E^{*}$. These two parameters completely determine the computational graph executed during inference.

Once the deployment configuration has been selected, the hybrid network is instantiated by assigning the first $K^{*}$ backbone stages to the neuromorphic SNN accelerator and the remaining stages to the ANN accelerator. The selected ANN early exits are inserted after the corresponding intermediate feature maps, while the learned controller makes the per-frame decision of whether inference terminates at an intermediate exit or continues toward the final detection head, with the confidence threshold serving as its baseline operating point. Since the SNN backbone is always executed completely, dynamic execution affects only the ANN component of the network.

Before deployment, the selective bit-level protection identified during the reliability characterization stage is applied to the ANN component. This protection is incorporated into every evaluated candidate to eliminate dominant hardware-induced failure mechanisms without modifying the network architecture or requiring retraining. Consequently, the final deployment preserves the original network parameters while improving robustness against transient hardware faults.

The deployed configuration therefore represents the optimal compromise among detection accuracy, energy consumption, execution time, and reliability for the selected optimization objective. Since the optimization framework is independent of the underlying hardware platform, different deployment priorities can be accommodated by modifying only the objective weights without changing the proposed methodology.

\section{Experimental Results}
\label{sec:results}

This section evaluates the proposed BLADE framework from four perspectives: the dynamic hybrid SNN--ANN architecture, the reliability characteristics of the hybrid detector, the reliability behavior across SNN--ANN boundaries, and the final reliability-aware boundary and ANN early-exit selection.

\subsection{Experimental Setup}
\label{sec:setup}

All experiments use the Prophesee Gen1 automotive event-camera object detection dataset~\cite{tournemire2020gen1}, recorded by a moving vehicle and annotated with two classes, cars and pedestrians. Following the base detector~\cite{ahmed2025efficient}, each event stream is accumulated into $T=10$ temporal bins at $240\times304$, and this timestep count is kept fixed in all experiments. Detection accuracy is reported as mAP@0.5.

Operating points are selected only on the validation split ($20\,329$ images) and re-evaluated on the held-out test split ($30\,605$ images) with paired bootstrap confidence intervals ($B=1000$, seed $0$). The deployed detector uses the hybrid boundary $K=4$. Cross-boundary reliability is evaluated for $K\in\{0,2,4,8\}$, from fully ANN ($K=0$) to fully SNN ($K=8$), in both floating-point and integer implementations.

Reliability characterization uses hierarchical statistical fault injection on a fixed subset of $256$ validation images, shared across all models and fault conditions for paired comparison. Hardware fault rates follow the technology-dependent BER anchors of Riera \textit{et al.}~\cite{riera2016ber}, spanning $5.59\times10^{-9}$ to $2.04\times10^{-5}$. This range defines the realistic operating region, while higher BER values are used only for stress analysis.

\subsection{Dynamic Hybrid SNN--ANN Architecture}
\label{sec:dynamic_exit}

This subsection evaluates the dynamic hybrid SNN--ANN architecture and selects the ANN early-exit operating point, exit location, and number of exits before the final reliability-aware BLADE optimization.

\subsubsection{ANN Early Exit}
\label{sec:ann_exit}

The proposed dynamic inference mechanism performs early termination only within the ANN component. After the ANN backbone, a lightweight routing head produces preliminary detections at the stride-16 feature level. The deployed detector uses a learned controller that maps a multi-dimensional detection state, formed by the top1--top2 score margin and the number of detected objects, to either exiting at the router or continuing to the full multi-scale detection head. This places the exit in the conditional-computation lineage of dynamic layer-skipping networks such as BlockDrop~\cite{wu2018blockdrop} and SkipNet~\cite{wang2018skipnet}, where a controller is trained by reinforcement learning~\cite{mnih2015dqn} to bypass parts of the network on a per-input basis.

A confidence threshold $\tau$ is used as the baseline and as a convenient notation for operating points on the energy--accuracy curve. Under this baseline, a scalar confidence score is computed from the router detections and compared with $\tau$. If the confidence exceeds the threshold, inference terminates at the ANN early exit; otherwise, execution continues to the final detection head.

\Cref{tab:dynamic_sweep} reports the energy--accuracy tradeoff on the Gen1 validation split. The static $K=4$ detector without early exit achieves $0.6789$ mAP@0.5 at $17.53$\,mJ, including the always-active routing logic. The always-exit configuration reduces energy to $10.85$\,mJ, corresponding to the maximum saving of $38.1\%$, but lowers mAP@0.5 to $0.5919$. The deployed operating point at $\tau=0.90$ achieves $0.6695$ mAP@0.5 and $16.22$\,mJ on the router-included operating-curve basis, reducing energy by $7.4\%$ with a $1.4\%$ relative mAP drop and exiting on $19.5\%$ of validation samples. At $\tau=0.92$, the exit rate falls to $7.39\%$, with $0.6761$ mAP@0.5 and $17.03$\,mJ.

\begin{table*}[t]
  \centering
  \caption{Energy--accuracy operating curve of the spatial neck router}
  \label{tab:dynamic_sweep}
  \begin{tabular}{lccccc}
    \toprule
    $\tau$ & Exit rate & mAP@0.5 $\uparrow$ & $\Delta$mAP (\%) & Energy (mJ) $\downarrow$ & Saved (\%) \\
    \midrule
    static ($\tau\!\ge\!0.96$) & 0.000 & 0.6789 & --- & 17.53 & 0.0 \\
    0.92 & 0.074 & 0.6761 & 0.4 & 17.03 & 2.8 \\
    \rowcolor{gray!15} 0.90 & 0.195 & 0.6695 & 1.4 & 16.22 & 7.4 \\
    0.88 & 0.312 & 0.6649 & 2.1 & 15.44 & 11.9 \\
    0.86 & 0.411 & 0.6638 & 2.2 & 14.79 & 15.6 \\
    0.84 & 0.491 & 0.6566 & 3.3 & 14.25 & 18.7 \\
    0.82 & 0.548 & 0.6551 & 3.5 & 13.87 & 20.9 \\
    0.80 & 0.593 & 0.6472 & 4.7 & 13.57 & 22.6 \\
    always-exit (router-only) & 1.000 & 0.5919 & 12.8 & 10.85 & 38.1 \\
    \bottomrule
  \end{tabular}
\end{table*}

\begin{figure}[t]
\centering
\includegraphics[width=\columnwidth]{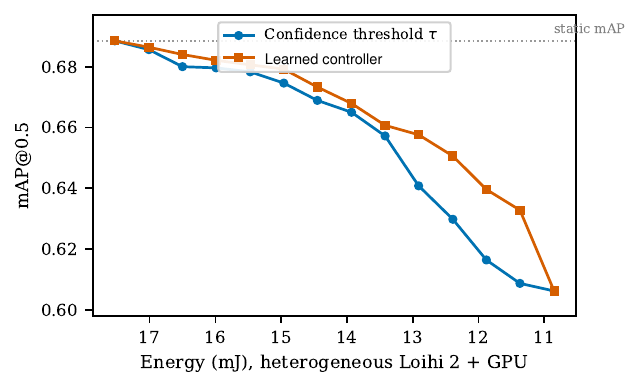}
\caption{Energy--accuracy tradeoff of the ANN early exit on the matched held-out validation half ($n{=}10{,}164$, seed~0): the learned controller (multi-dimensional detection state) versus the confidence-threshold ($\tau$) baseline.}
\label{fig:c1_pareto}
\end{figure}

\Cref{fig:c1_pareto} shows the energy--accuracy frontier. Lower thresholds increase the number of exited samples and reduce energy at the cost of accuracy, while higher thresholds approach the static detector. Since the complete SNN backbone and ANN feature extraction are always executed before the exit, only the final ANN detection head is skipped. Therefore, the dynamic saving is structurally capped at $38.1\%$.

The exit primarily saves energy rather than latency. On the evaluated GPU platform, the SNN backbone accounts for about $96\%$ of total inference latency, so skipping the ANN detection head reduces latency by only about $3.1\%$ despite the larger energy saving. The validation-selected thresholds are also evaluated on the held-out test split. The static detector reaches $0.6679$ mAP@0.5 with a $95\%$ confidence interval of $[0.6572,0.6801]$. At $\tau=0.90$, the early exit introduces a $0.45$ percentage-point mAP drop at a $19.32\%$ test exit rate. At $\tau=0.92$, the drop decreases to $0.15$ percentage points, remaining within the bootstrap confidence interval.

The learned controller improves over the confidence threshold when it observes the full detection state. A tabular controller restricted to binned scalar confidence collapses to a near-never-exit policy, retaining $0.6788$ mAP@0.5, exiting only $0.0075$ of samples, and saving $0.3\%$ energy (\Cref{tab:sparq_ablations}). This confirms that scalar confidence alone provides no additional information beyond the threshold baseline.

\begin{table*}[t]
  \centering
  \caption{Controller ablation on the dynamic early-exit hybrid}
  \label{tab:sparq_ablations}
  \begin{tabular}{lcccc}
    \toprule
    Controller & mAP@0.5 $\uparrow$ & Exit rate & Energy (mJ) $\downarrow$ & Saved (\%) \\
    \midrule
    \rowcolor{gray!15} Static (no exit) & 0.6789 & 0.0000 & 17.53 & 0.0 \\
    Binned-confidence tabular-$Q$ (scalar-confidence state) & 0.6788 & 0.0075 & 17.48 & 0.3 \\
    Occupancy gate $n_{\text{det}}\!\le\!0$ (1 line) & 0.6763 & 0.0551 & 17.16 & 2.1 \\
    Threshold $\tau{=}0.80$ & 0.6472 & 0.5933 & 13.57 & 22.6 \\
    \bottomrule
  \end{tabular}
\end{table*}

In contrast, a learned controller over the continuous multi-dimensional state matches or improves the threshold at every non-trivial operating point. \Cref{tab:c1_multiseed} reports the five-seed matched-exit comparison on the held-out split. The paired-$t$ gain interval ($\mathrm{df}=4$) excludes zero at $10$ of $12$ non-trivial exit fractions, with peak gain $+0.0326\pm0.0062$ mAP@0.5 at $80\%$ exit, $+0.0278\pm0.0060$ at $72\%$, and $+0.0207\pm0.0059$ at $64\%$. At low exit rates the difference is insignificant, for example $-0.0020$ at $8\%$. The learned controller therefore dominates the threshold when energy savings are large and matches it when savings are small.

\begin{table}[t]
  \centering
  \caption{Five-seed significance of the learned exit versus the confidence threshold}
  \label{tab:c1_multiseed}
  \begin{tabular}{ccccc}
    \toprule
    Exit frac & E (mJ) & $\tau$ mAP $\uparrow$ & Learned mAP $\uparrow$ & $\Delta$(Learned$-\tau$) \\
    \midrule
    0.08 & 17.0 & 0.6606 & 0.6586 & $-0.0020\!\pm\!0.0021$ \\
    0.16 & 16.5 & 0.6532 & 0.6532 & $-0.0000\!\pm\!0.0039$ \\
    0.24 & 15.9 & 0.6455 & 0.6494 & $+0.0039\!\pm\!0.0032$$^\dagger$ \\
    0.32 & 15.4 & 0.6408 & 0.6470 & $+0.0061\!\pm\!0.0042$$^\dagger$ \\
    0.40 & 14.9 & 0.6327 & 0.6432 & $+0.0105\!\pm\!0.0041$$^\dagger$ \\
    0.48 & 14.3 & 0.6252 & 0.6361 & $+0.0108\!\pm\!0.0033$$^\dagger$ \\
    0.56 & 13.8 & 0.6160 & 0.6278 & $+0.0118\!\pm\!0.0047$$^\dagger$ \\
    0.64 & 13.3 & 0.5965 & 0.6173 & $+0.0207\!\pm\!0.0059$$^\dagger$ \\
    0.72 & 12.7 & 0.5790 & 0.6068 & $+0.0278\!\pm\!0.0060$$^\dagger$ \\
    0.80 & 12.2 & 0.5591 & 0.5917 & $+0.0326\!\pm\!0.0062$$^\dagger$ \\
    0.88 & 11.7 & 0.5430 & 0.5733 & $+0.0303\!\pm\!0.0061$$^\dagger$ \\
    0.96 & 11.1 & 0.5310 & 0.5526 & $+0.0216\!\pm\!0.0027$$^\dagger$ \\
    \midrule
    \multicolumn{5}{@{}l@{}}{\footnotesize $\dagger$ paired-$t$ interval (df${=}4$) excludes zero.} \\
    \bottomrule
  \end{tabular}
\end{table}

The improvement mainly comes from the multi-dimensional detection state rather than the reinforcement-learning formulation itself. A non-reinforcement linear gate using the same standardized state features improves over the threshold at $11$ of $12$ fractions and recovers approximately $86\%$ of the best gain achieved by the learned controller. This is consistent with the weak correlation between scalar confidence and true exit loss, measured at only $-0.108$.

\subsubsection{Exit Placement}
\label{sec:exit_placement}

The ANN early exit is evaluated at the three feature pyramid levels, stride-8 ($p3$), stride-16 ($p4$), and stride-32 ($p5$). Since all configurations share the same backbone, the static detector is identical and only the exit behavior changes.

The always-exit configuration shows how much information each feature level preserves when the final head is skipped. The stride-16 exit achieves the highest mAP@0.5, $0.5781$, compared with $0.4510$ at stride-32 and $0.4080$ at stride-8. Under the deployment-relevant near-lossless operating point, stride-16 again performs best, reaching $0.6719$ mAP@0.5 at a $9.7\%$ exit rate, ahead of stride-32 with $0.6529$ at $8.7\%$ and stride-8 with $0.6217$ at $17.8\%$.

For completeness, stride-16 and stride-32 are also compared at a common exit fraction of $13.7\%$. Under this artificial equalization, stride-32 reaches $0.659$ mAP@0.5 and stride-16 reaches $0.655$, while stride-8 reaches comparable accuracy only at a much lower exit fraction. Since deployment uses the near-lossless operating criterion rather than forced equal exit rates, the stride-16 feature level ($p4$) is selected for the ANN early exit.

\subsubsection{Number of ANN Exits}
\label{sec:number_exits}

The final architectural choice is the number of exits. First, a temporal exit before the SNN--ANN boundary is evaluated to test whether inference can terminate during SNN execution. The temporal gate is inserted before the ASAB bridge and evaluated over $\alpha\in\{0.85,0.90,0.95,0.99\}$ and $\tau\in\{0.10,0.25,0.40,0.55,0.70,0.99\}$. Across the entire grid, the temporal exit never fires, giving a temporal exit rate of $0.000$ and an average exit timestep of $10$ out of $10$. This occurs because the ASAB bridge requires all $T=10$ temporal bins before producing the fused ANN feature representation, so no reliable pre-fusion decision is available.

A second ANN spatial exit is then evaluated by cascading a stride-8 ($p3$) exit before the deployed stride-16 ($p4$) exit. At the accuracy-matched operating point, the first exit fires on $0\%$ of frames, so the cascade collapses to the single $p4$ exit. The dual-exit configuration reaches the same $0.6722$ mAP@0.5 at an overall exit rate of about $17\%$, and across the evaluated fault rates follows the same graceful-degradation behavior without catastrophic failures in the realistic region.

The additional exit only increases overhead. In the dual-exit study, two always-active single-tap routers consume about $1.25$\,mJ, compared with $0.25$\,mJ for the corresponding single-tap ANN exit, a $5\times$ increase without accuracy or reliability benefit. These single-tap values are distinct from the deployed multi-scale router ($\approx1.31$\,mJ), which reads $p3$, $p4$, and $p5$. The final architecture therefore uses one ANN early exit at stride-16.

\subsection{Reliability Characterization}
\label{sec:reliability}

This subsection characterizes the reliability of the dynamic hybrid SNN--ANN detector. The analysis first identifies the dominant vulnerable bit, then localizes the affected layers, then evaluates the protected detector under multi-bit BER faults, and finally compares reliability across SNN--ANN boundaries.

\subsubsection{Bit-level Vulnerability Analysis}
\label{sec:bit_level}

A single-bit fault-injection campaign is first performed without protection to identify the dominant fault mechanism. The results reveal a highly localized floating-point vulnerability, consistent with prior observations that bit-level sensitivity is nonuniform~\cite{yang2026memory}. The most significant exponent bit, bit~30, dominates the ANN fault behavior, while almost all other bit positions are benign.

Corrupting bit~30 produces significant-or-worse accuracy degradation in $58.8\%$ of injections, with Wilson $95\%$ confidence interval $[55.6,61.9]\%$, corresponding to $544$ vulnerable injections out of $925$. In contrast, the remaining $31$ floating-point bit positions jointly produce only one significant degradation over $3155$ injections, with Wilson upper bound $0.18\%$. The cause is the IEEE-754 exponent field: flipping bit~30 changes the exponent bias by $128$ and rescales the affected weight by approximately $2^{128}$ or $2^{-128}$. The int8 SNN backbone has no equivalent exponent bit and therefore degrades gracefully without catastrophic failures over the evaluated fault range.

Since the vulnerable bit is fixed and known, one parity or error-correcting-code bit~\cite{hamming1950} per floating-point weight is sufficient to protect it, corresponding to approximately $3\%$ storage overhead. This protection is applied in the remaining reliability experiments.

\begin{table}[t]
\centering
\caption{Single-bit catastrophe census by IEEE-754 field, fp32 weights}
\label{tab:bitfield}
\begin{tabular}{lccc}
\toprule
IEEE-754 field & Bits & Injections & Catastrophe rate \\
\midrule
Sign            & 31      & 25   & 0.00\% \\
Exponent MSB    & 30      & 925  & 58.81\% \\
Exponent (rest) & 23--29  & 2555 & 0.04\% \\
Mantissa        & 0--22   & 575  & 0.00\% \\
\bottomrule
\end{tabular}
\end{table}

\subsubsection{Layer-wise Criticality}
\label{sec:layerwise}

\begin{figure}[t]
\centering
\includegraphics[width=\columnwidth]{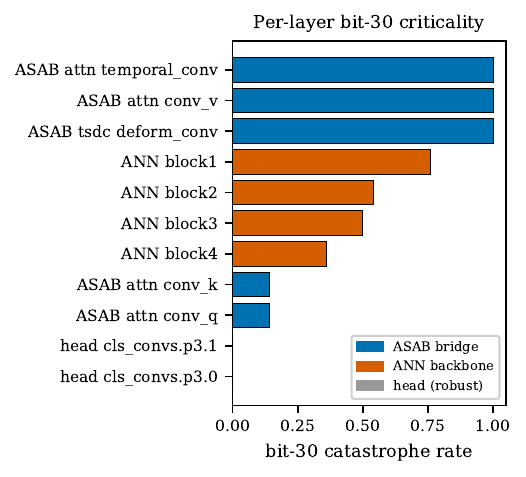}
\caption{Layer-wise catastrophic failure rate under exponent-MSB (bit~30) fault injection, concentrated in the ASAB bridge and early ANN backbone.}
\label{fig:layer_criticality}
\end{figure}

A layer-wise campaign then injects bit~30 faults into each floating-point layer independently. \Cref{fig:layer_criticality} shows that vulnerability is concentrated immediately after the SNN--ANN boundary. The ASAB bridge is catastrophic in every evaluated trial, with catastrophic fraction $1.0$, and the first ANN backbone block follows with catastrophic fraction $0.76$. The vulnerability then decreases through later ANN blocks.

The ANN detection head is robust despite containing the largest number of floating-point parameters. Every evaluated detection-head layer records catastrophic fraction zero. Overall, about $49\%$ of floating-point parameters belong to the vulnerable ASAB bridge and ANN backbone, while the remaining $51\%$ reside mostly in the robust detection head. Selective protection can therefore focus on the measured critical layers, and applying bit-level protection eliminates catastrophic behavior across all evaluated layers.

\subsubsection{Multi-bit Reliability}
\label{sec:multibit}

\begin{figure}[t]
\centering
\includegraphics[width=\columnwidth]{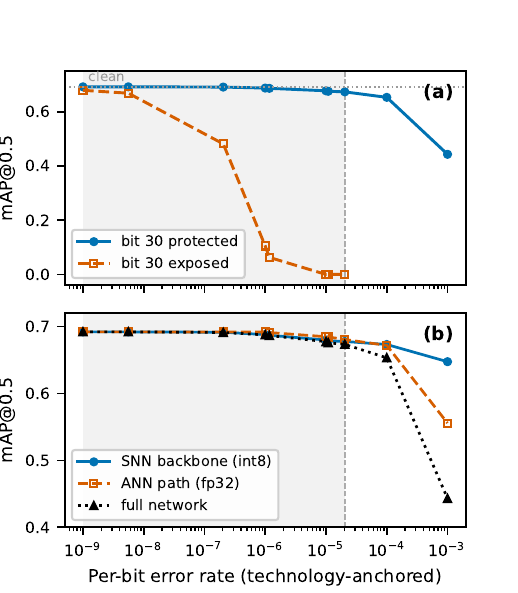}
\caption{Reliability of the protected detector under increasing BER. (a)~Full network, before versus after bit-level protection. (b)~Protected detector by substrate (SNN, ANN, full network).}
\label{fig:map_vs_ber}
\end{figure}

After bit-level protection, the deployed hybrid detector is evaluated under multi-bit BER faults. For each BER, faults are injected independently into network weights, and the detector is evaluated over $M=100$ Monte Carlo trials on the fixed $256$-image subset. \Cref{tab:multibit_ber} summarizes the results, and \Cref{fig:map_vs_ber} shows the degradation trend.

Protecting bit~30 removes catastrophic failures throughout the realistic technology range. The detector instead degrades gracefully: mean mAP@0.5 decreases from $0.692$ clean to $0.686$ at $10^{-5}$, $0.677$ at $10^{-4}$, $0.668$ at $10^{-3}$, $0.572$ at $10^{-2}$, and $0.194$ at the extreme stress point $5\times10^{-2}$. The fail fraction follows the same trend, remaining zero throughout the realistic region and increasing only at higher BER values.

The two substrates degrade differently. The int8 SNN backbone begins to lose accuracy at lower rates but degrades more gradually, while the protected fp32 ANN component retains higher accuracy at moderate BER values before dropping more sharply at extreme fault rates. No catastrophic trial is observed for either substrate within the realistic BER range from $5.59\times10^{-9}$ to $2.04\times10^{-5}$~\cite{riera2016ber}.

The result is robust to evaluation subset and sampling size. Repeating the protected campaign on independent subsets preserves the degradation trend. Increasing the Monte Carlo count from $100$ to $1000$ tightens the Wilson confidence interval without changing the reliability behavior, and repeating the campaign on a larger $2048$-image subset confirms that the graceful degradation is not an artifact of the fixed $256$-image subset.

\begin{table}[t]
\centering
\caption{Per-bit soft-error stress anchors from the Riera 6T-SRAM characterisation~\cite{riera2016ber}}
\label{tab:riera_anchors}
\begin{tabular}{lcc}
\toprule
Technology & Node (nm) & SER (FIT/cell)~\cite{riera2016ber} \\
\midrule
Bulk planar & 22 & $2.04\times10^{-5}$ \\
Bulk planar & 16 & $1.09\times10^{-5}$ \\
SOI planar  & 22 & $1.17\times10^{-6}$ \\
Bulk FinFET & 20 & $2.04\times10^{-7}$ \\
Bulk FinFET & 14 & $5.59\times10^{-9}$ \\
\bottomrule
\end{tabular}
\end{table}

\begin{table*}[t]
\centering
\caption{Multi-bit per-weight BER sweep of the deployed $K{=}4$ hybrid (bit~30 protected)}
\label{tab:multibit_ber}
\begin{tabular}{lccccccc}
\toprule
 & clean & $10^{-6}$ & $10^{-5}$ & $10^{-4}$ & $10^{-3}$ & $10^{-2}$ & $5\!\times\!10^{-2}$ \\
\midrule
\multicolumn{8}{l}{\emph{Mean mAP@0.5}}\\
\quad SNN backbone (int8) & 0.692 & 0.692 & 0.687 & 0.679 & 0.673 & 0.648 & 0.524 \\
\quad ANN path (fp32)     & 0.692 & 0.692 & 0.692 & 0.688 & 0.680 & 0.650 & 0.468 \\
\quad Full network        & 0.692 & 0.691 & 0.686 & 0.677 & 0.668 & 0.572 & 0.194 \\
\midrule
\multicolumn{8}{l}{\emph{Full network, fault impact}}\\
\quad Fail fraction ($\ge5\%$ drop) & 0 & 0 & 0 & 0.01 & 0.08 & 0.92 & 1.00 \\
\quad Mean weights flipped          & -- & 7 & 66 & 662 & 6.6k & 66k & 332k \\
\bottomrule
\end{tabular}
\end{table*}

\subsubsection{Cross-boundary Reliability}
\label{sec:cross_boundary}

The protected reliability analysis is extended to the boundary family $K\in\{0,2,4,8\}$. All configurations use the same bit-level protection. \Cref{tab:crossmodel_ber} reports the cross-boundary reliability results.

All boundaries degrade gracefully across the realistic technology range, and none produces catastrophic failures after protecting the floating-point exponent MSB. Therefore, the reliability difference between boundaries appears mainly as retained detection accuracy rather than catastrophic failure probability. Increasing the SNN portion improves robustness under aggressive fault conditions, with the fully SNN configuration ($K=8$) retaining the highest accuracy across the stress range, while the deployed hybrid configuration ($K=4$) provides the highest clean detection accuracy.

This confirms the core BLADE tradeoff: more SNN computation improves robustness because less floating-point computation remains exposed, but the same shift can reduce clean accuracy. Within the realistic technology region, every protected configuration remains catastrophe-free and close to its clean baseline, so reliability distinguishes boundaries mainly under increasing hardware stress rather than as an in-band survival constraint.

\begin{table*}[t]
\centering
\caption{Cross-model clean accuracy and catastrophic fraction (bit~30 protected)}
\label{tab:crossmodel_ber}
\begin{tabular}{lcc}
\toprule
Model & clean mAP@0.5 $\uparrow$ & Catastrophic frac.\ @ $5{\times}10^{-2}$ \\
\midrule
Fully ANN $K{=}0$, unquantized & 0.652 & 1\% \\
Fully ANN $K{=}0$, quantized & 0.662 & 0\% \\
Hybrid $K{=}2$, unquantized & 0.663 & 20\% \\
Hybrid $K{=}2$, quantized & 0.669 & 4\% \\
\rowcolor{gray!15} Hybrid $K{=}4$ (deployed), unquantized & 0.692 & 4\% \\
\rowcolor{gray!15} Hybrid $K{=}4$, fully int8 (head quantized; counterfactual) & 0.691 & 23\% \\
Fully SNN $K{=}8$, unquantized & 0.661 & 0\% \\
Fully SNN $K{=}8$, quantized & 0.646 & 0\% \\
\bottomrule
\end{tabular}
\end{table*}

\subsection{Reliability-aware Boundary Selection}
\label{sec:blade_results}

The final experiment evaluates BLADE by jointly optimizing the SNN--ANN boundary and ANN early-exit configuration according to reliability, detection accuracy, execution time, and energy consumption. Each candidate is evaluated using the objective formulation of Section~\ref{sec:optimization}. Reliability is the retained detection accuracy under the complete BER sweep, while the remaining objectives come from the experimental characterization.

\Cref{tab:blade} reports the selected deployment configurations. When reliability and energy dominate, BLADE selects the fully SNN configuration ($K=8$) with the stride-16 ANN early exit, achieving the highest reliability retention of $0.965$ and reducing inference energy to $10.20$\,mJ. When detection accuracy dominates, BLADE selects the deployed hybrid configuration ($K=4$), which achieves the highest mAP@0.5 of $0.691$ and requires $15.82$\,mJ per inference on the compute-only basis used in \Cref{tab:blade}, where the stride-16 exit fires and the always-on router energy is excluded.

Although the $K=4$ configuration is less robust under extreme fault conditions, it remains free of catastrophic failures throughout the realistic technology range after bit-level protection. The results show that no single boundary optimizes all objectives simultaneously; the selected deployment depends on the target balance among reliability, accuracy, execution time, and energy consumption.

\begin{table}[t]
\centering
\caption{\textsc{BLADE} feasible designs ranked by balanced score ($\star$ reliability--energy optimum, $\Diamond$ deployed accuracy-priority)}
\label{tab:blade}
\setlength{\tabcolsep}{4pt}
\resizebox{\columnwidth}{!}{%
\begin{tabular}{lccccc}
\toprule
$K$ & Exit & Retention & mAP@0.5 & \shortstack{Energy\\(mJ)} & Score \\
\midrule
8 & $p4$   & 0.965 & 0.660 & 10.20 & $0.672\,\star$ \\
8 & none   & 0.965 & 0.661 & 10.60 & 0.555 \\
4 & $p3$   & 0.914 & 0.682 & 16.03 & 0.432 \\
4 & $p4$   & 0.914 & 0.691 & 15.82 & $0.417\,\Diamond$ \\
4 & $p5$   & 0.914 & 0.690 & 15.64 & 0.400 \\
4 & none   & 0.914 & 0.692 & 16.22 & 0.300 \\
\midrule
\multicolumn{6}{@{}p{0.95\columnwidth}@{}}{\footnotesize Retention $=$ mean mAP@0.5 held across the BER ladder. Energy is operation-count compute-only, excluding the $1.31$\,mJ always-on router; rows with an exit give the energy when that exit fires (e.g.\ $15.82$\,mJ at $K{=}4$/$p4$ versus $16.22$\,mJ never-exit).} \\
\bottomrule
\end{tabular}}
\end{table}

Overall, reliability-aware boundary selection changes the preferred deployment compared with accuracy- or energy-only optimization. The fully SNN implementation provides the strongest robustness under aggressive faults, while the deployed hybrid configuration provides the best detection accuracy and remains catastrophe-free throughout the realistic technology range after selective protection.

\section{Conclusion}
\label{sec:conclusion}

This paper presented \textsc{BLADE}, the first reliability-aware boundary selection methodology for dynamic hybrid SNN--ANN networks with ANN early exit. The proposed framework jointly optimizes the SNN--ANN boundary and ANN early-exit configuration according to reliability, detection accuracy, execution time, and energy consumption, enabling deployment configurations tailored to different application requirements.

Experimental results demonstrate that the selected deployment achieves an mAP@0.5 of $0.691$ with an inference compute energy of $15.82$~mJ when the ANN early exit fires, while a fully SNN implementation reduces the energy consumption to $10.20$~mJ when reliability and efficiency are prioritized. The proposed reliability characterization further identifies a single floating-point exponent bit as the dominant source of failures, with bit-30 corruption causing significant-or-worse degradation in $58.8\%$ of injections. Protecting this bit with only approximately $3\%$ storage overhead eliminates catastrophic failures across all evaluated realistic technology fault rates, while the fully SNN configuration achieves the highest reliability retention of $0.965$ under aggressive fault conditions. These results demonstrate that reliability-aware boundary selection enables more dependable hybrid SNN--ANN deployments without sacrificing competitive detection accuracy and energy efficiency.

The proposed methodology requires only candidate SNN--ANN boundaries, ANN early exits, and per-candidate accuracy, energy, execution time, and reliability metrics, making it applicable to a broad range of hybrid SNN--ANN architectures. Future work will extend the methodology to additional event-based perception tasks, activation and permanent fault models, and experimental validation on emerging neuromorphic hardware platforms.

\section*{ACKNOWLEDGMENT}
\small
This work was supported in part by the Estonian Research Council grant PUT PRG1467 ``CRASHLESS'', EU Grant Project 101160182 ``TAICHIP'', and by the Federal Ministry of Research, Technology and Space of Germany (BMFTR) for supporting Edge-Cloud AI for DIstributed Sensing and COmputing (AI-DISCO) project (Project-ID ``16ME1127'').
\bibliographystyle{IEEEtran}
\bibliography{references}

\end{document}